%% file: main.tex
\documentclass[letterpaper]{article} 
\usepackage{aaai2026}  
\usepackage{times}  
\usepackage{helvet}  
\usepackage{courier}  
\usepackage[hyphens]{url}  
\usepackage{graphicx} 
\usepackage{natbib}  
\usepackage{caption} 
\usepackage{algorithm}
\usepackage{algorithmic}

\usepackage{newfloat}
\usepackage{listings}
\DeclareCaptionStyle{ruled}{labelfont=normalfont,labelsep=colon,strut=off} 
\floatstyle{ruled}
\newfloat{listing}{tb}{lst}{}
\floatname{listing}{Listing}
\usepackage{amsmath}
\usepackage{amsfonts}
\usepackage{multirow}
\usepackage{tabularx}
\usepackage{booktabs} 
\usepackage{xspace}
\usepackage[capitalize]{cleveref}
\usepackage{xcolor}
\usepackage{soul}
\definecolor{myblue}{RGB}{0, 102, 204}
\newcommand{\SD}{\textsc{SD\textsuperscript{2}}\xspace}

\newif\ifisextended
\isextendedtrue      

\newcommand{\appref}[1]{%
  \ifisextended
    \cref{#1}%
  \else
    the extended version%
  \fi
}

\ifisextended
    \nocopyright
\fi

\title{Steering Pretrained Drafters during Speculative Decoding}
\author{
    Frédéric Berdoz,
    Peer Rheinboldt,
    Roger Wattenhofer
}

\affiliations{
    ETH Zurich\\
    \{fberdoz, prheinboldt, wattenhofer\}@ethz.ch
}

\begin{document}

\maketitle

\ifisextended
    \insert\footins{\noindent\footnotesize This is the extended version of the corresponding paper published in the AAAI 2026 proceedings.}
\fi

\begin{abstract}
Speculative decoding accelerates language model inference by separating generation into fast drafting and parallel verification. Its main limitation is drafter–verifier misalignment, which limits token acceptance and reduces overall effectiveness.  While small drafting heads trained from scratch compensate with speed, they struggle when verification dominates latency or when inputs are out of distribution. In contrast, pretrained drafters, though slower, achieve higher acceptance rates thanks to stronger standalone generation capabilities, making them competitive when drafting latency is negligible relative to verification or communication overhead. In this work, we aim to improve the acceptance rates of pretrained drafters by introducing a lightweight dynamic alignment mechanism: a \emph{steering vector} computed from the verifier’s hidden states and injected into the pretrained drafter. Compared to existing offline alignment methods such as distillation, our approach boosts the number of accepted tokens by up to 35\% under standard sampling and 22\% under greedy sampling, all while incurring negligible computational overhead. Importantly, our approach can be retrofitted to existing architectures and pretrained models, enabling rapid adoption.
\end{abstract} 

\begin{links}
    \link{Code}{https://github.com/ETH-DISCO/SD-square}
    \ifisextended \else
    \link{Extended version}{https://arxiv.org/abs/XXXXXX} 
    \fi
\end{links}

\section{Introduction}
The auto-regressive nature of transformer-based large language models (LLMs) \cite{vaswani2017attention} inherently limits their inference speed. This limitation is further amplified by the rapid growth in model size among frontier LLMs \cite{achiam2023gpt, grattafiori2024llama, liu2024deepseek, yang2025qwen3}. Numerous approaches have been proposed to reduce latency, including weight quantization \cite{dettmers2022gpt3}, model pruning \cite{han2016deep}, and distillation \cite{hinton2015distilling}, but these often come at the expense of generated text quality. A paradigm that escapes this trade-off is \emph{speculative decoding} \cite{leviathan2023fast, xia2023speculative, chen2023accelerating}, which follows the general principle of \emph{speculative execution} \cite{burton2012speculative}. This method employs a lightweight \emph{drafter} to propose the next $k$ tokens, which are then verified in parallel using a single forward pass of the larger base model, commonly referred to as the \emph{verifier}. In essence, speculative decoding leverages the underutilization of accelerator hardware in classic auto-regressive decoding by using batched verification to amortize the costly transfer of model parameters between off-chip memory and on-chip cache.
Two main families of approaches have emerged for speculative decoding \cite{hu2025speculative}. The first uses an independent drafter \cite{xia2023speculative}, typically a compact LLM trained independently on similar data as the verifier. Since these drafters are capable language models in their own right, they can generalize reasonably well, even without task-specific tuning or dynamic steering.
The second family of approach employs small dependent speculative heads mounted directly on top of the verifier and trained from scratch \cite{stern2018blockwise, cai2024medusa, ankner2024hydra, li2024eagle1}. 
At inference, these methods rely mostly on dynamic steering to keep the drafter aligned with the verifier despite its limited capacity. Although such drafters often produce shorter accepted blocks, their low latency allows them to rapidly generate many candidate sequences. Combined with efficient batch evaluation \cite{miao2024specinfer}, this makes them competitive in settings where the cost of verification is relatively low, such as in controlled research environments. However, in real-world scenarios where verification latency fluctuates or dominates total runtime, e.g., when the verifier is remote \cite{openai2024predicted}, deployed on slower hardware, shared across several drafters, or simply frontier-scale with 600B+ parameters \cite{liu2024deepseek}, the block efficiency becomes the key driver of efficacy, as it dictates the number of verification steps. 
While independent drafters tend to perform better in that regard, due to their ability to generate coherent sequences, they can only rely on their offline alignment with the verifier. Building on the observation that LLMs (verifiers in our case) implicitly encode information about upcoming tokens in their intermediate representations \cite{samragh2025your}, we propose \textbf{S}teering pretrained \textbf{D}rafters during \textbf{S}peculative \textbf{D}ecoding (\SD), a lightweight guiding mechanism that extracts this latent signal to dynamically steer drafters at inference.
\begin{figure*}[t]
    \centering
    \includegraphics[width=\textwidth]{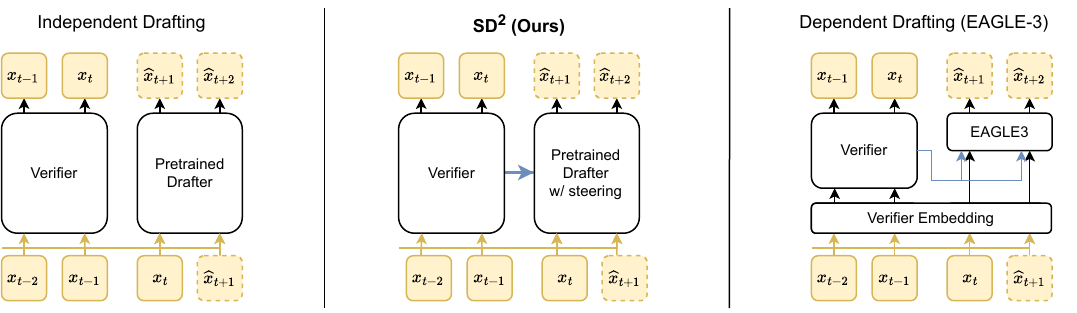}
    \caption{Overview of different drafting paradigms: Independent drafting uses a smaller model from the same family as the verifier, with no access to its internal state. Dependent drafting (e.g., EAGLE-3) uses lightweight heads trained to read the verifier’s hidden states, sharing input embeddings and using concatenated features for guidance. \SD strikes a middle ground, leveraging verifier features for steering while retaining the generalization capabilities of independent drafters.}
    \label{fig:comparison}
\end{figure*}

Our key contributions include:
\begin{itemize}
    \item We introduce a lightweight dynamic steering mechanism for pretrained drafters during speculative decoding.
    \item We show that our steering mechanism improves the number of drafted tokens accepted by up to 35\% and has up to 22\% higher throughput compared to independent drafters across a variety of tasks and models.
    \item We motivate our design choices with several ablations.
\end{itemize}

\section{Related Work}

\subsection{Speculative Decoding}
Speculative decoding (SD) originates from the speculative execution paradigm \cite{burton2012speculative}. While early variants only supported greedy decoding acceleration \cite{stern2018blockwise,sun2021instantaneous,ge2022lossless,xia2023speculative}, the concurrent works of \citet{leviathan2023fast} and \citet{chen2023accelerating} introduced \emph{speculative sampling}, extending speculative decoding to non-deterministic decoding algorithms. 
This sparked a long line of work focused on improving the efficiency of such methods, typically evaluated by token throughput (wall-clock speedup) in controlled environments. We refer to \citet{hu2025speculative} for a comprehensive survey and detailed taxonomy of speculative decoding.

\paragraph{Dependent Drafters.}
The first drafters consisted of several decoding heads that independently drafted tokens to form a sequence, taking the verifier's last hidden state as input \cite{stern2018blockwise, cai2024medusa}. While fast (thanks to parallel token drafting), these methods suffer from the lack of dependency between the drafted tokens, strongly limiting the token acceptance rate. Recognizing this limitation, \citet{li2024eagle1} propose to use auto-regressive drafters on the hidden states, and \citet{ankner2024hydra} improves by taking the embeddings of the previously drafted tokens as input to the autoregressive drafter. Instead of only using the last hidden representations of the drafter, \citet{zimmer2025mixture} and \citet{du2024glide} use the KV values of the verifiers during drafting. \citet{zhang2025learning} and \citet{li2025eagle3} further improve the acceptance rates by training the drafter to use its hidden features to close the gap between training and inference. Although our study centers on independent drafters, we also report the block efficiency of EAGLE-3 \cite{li2025eagle3} in a chain decoding setting (i.e., only one proposed sequence, excluding its tree decoding component) to provide a reference point for the improvements achievable by independent drafters.

\paragraph{Independent Drafters.}
Independent auto-regressive drafters were first introduced by \citet{xia2023speculative}. Building on this, \citet{huang2024specdec} proposed an enhanced version where the candidate length is determined on the fly via an acceptance prediction head. \citet{zhou2024distillspec} note that the acceptance rate of the drafted token is theoretically bounded by the divergence between the drafter and verifier, and therefore propose to distill the verifier into the drafter. Alternatively, \cite{liu2024online} propose online speculative decoding, where drafters are continuously retrained on new user inputs, and \citet{fu2024break} propose a drafter-free version using intermediate Jacobi iterations as drafted sequences.

\paragraph{Verification.}
\citet{sun2023spectr,sun2024block} frame the verification phase as an optimal transport problem to improve batch and block verification, respectively. \citet{spector2023accelerating} and \citet{miao2024specinfer} introduce tree-based speculative inference, where many drafted sequences are arranged in a tree and verified in parallel. Building on this idea, \citet{li2024eagle2} introduce dynamic drafting trees. Lastly, \cite{yin2024theoretical} explore the theoretical limits of speculative decoding.

\subsection{Dynamic Steering of LLMs}
The technique of activation steering, first proposed by \citet{turner2023steering}, allows for the control of LLM behavior by directly modifying model activations during inference. It is primarily motivated by the \emph{linear representation hypothesis} \cite{park2024linear}, suggesting that a model’s intent or behavior is encoded along specific, steerable directions. Subsequently, \citet{rimsky2024steering} introduced a method to compute steering vectors by averaging the activation differences between sets of positive and negative examples. More recently, \citet{chalnev2024improving} introduce a method to predict a steering vector's impact on internal sparse autoencoder (SAE) features \cite{huben2024sparse}. However, these approaches focus on static, interpretable steering and remain largely unexplored in the dynamic context of speculative decoding.

\section{Methodology}

\begin{figure}[t]
    \centering
    \includegraphics[width=0.9\linewidth]{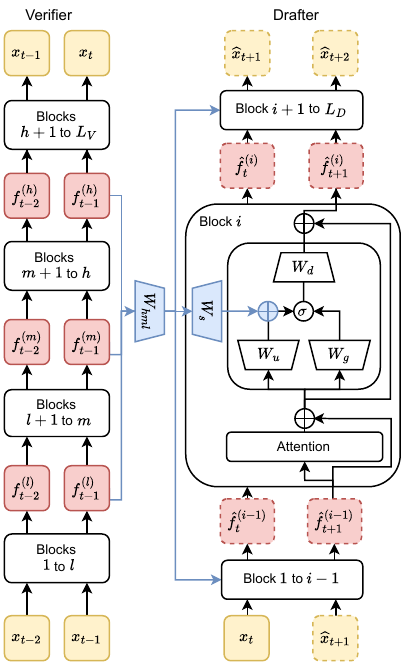}
    \caption{The steering mechanism in \SD works by concatenating the verifier’s high-, medium-, and low-level hidden features and passing them through a linear projection to produce a steering vector. This embedding is transformed by another linear layer into a set of biases, which are added to all MLP hidden states in the drafter just before the activation function, as detailed in \cref{eq:before_steering} and \cref{eq:after_steering}.}
    \label{fig:overview}
\end{figure}

\begin{figure}[t]
    \centering
    \includegraphics[width=0.9\linewidth]{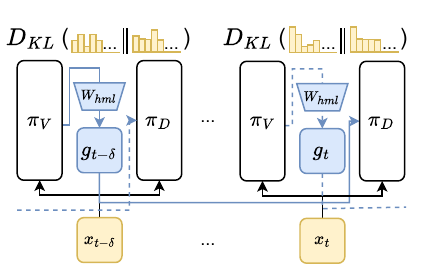}
    \caption{
The training process of \SD aligns the drafter's ($\pi_D$) probability distribution to the verifier's ($\pi_V$). To achieve this, we randomly choose an offset $\delta \in [1,k]$ to simulate drafting the $\delta$'th token of a block. After extracting $g$ from on the verifier's activations, we compute $\pi_D(x_t | x_{1:t-1}, g_{t-\delta})$ and use the Kullback-Leibler divergence $D_{\text{KL}}(\pi_V(\cdot |x_{1:t-1}) \Vert \pi_D(\cdot |x_{1:t-1}, g_{t-\delta}))$ as loss. In addition to $W_{s}$ (see~\cref{fig:overview}), both $W_{hml}$
and $\pi_D$ are trained. The verifier $\pi_V$ stays frozen throughout training. }
    \label{fig:training}
\end{figure}
Steered speculative decoding (\SD) follows the standard speculative decoding paradigm of drafting a candidate sequence and verifying each token in parallel \cite{leviathan2023fast, chen2023accelerating}. The key addition is that, in addition to the rejection of candidate tokens, the verification step also produces a \emph{steering vector}, which is used to guide the drafter in the next generation phase. Our method is motivated by the observation that auto-regressive models implicitly encode information about future tokens beyond the immediate next token, even without being explicitly trained to do so \cite{samragh2025your}. We aim to extract this predictive information from the verifier’s hidden representations and inject it into the drafter to dynamically guide generation.

\newcommand{\vdSize}[0]{0.8in}
\setlength{\tabcolsep}{1.2mm}
\begin{table*}[!ht]  
\centering 
\small 

\begin{tabularx}{\linewidth}{lX cccccccccccccc} \toprule
\textbf{Verifier \&} & \textbf{Method} & \multicolumn{2}{c}{\textbf{UltraChat}} & \multicolumn{2}{c}{\textbf{HumanEval}} & \multicolumn{2}{c}{\textbf{XSum}} & \multicolumn{2}{c}{\textbf{Alpaca}} & \multicolumn{2}{c}{\textbf{GSM8K}} & \multicolumn{2}{c}{\textbf{Mean}}\\ 
 \textbf{Drafter} & & $\tau$ & $\alpha$ & $\tau$ & $\alpha$ & $\tau$ & $\alpha$ & $\tau$ & $\alpha$ & $\tau$ & $\alpha$ & $\tau$ & $\alpha$ \\

\midrule & \multicolumn{12}{c}{T=1 (Sampling)} \\
\midrule 
\multirow{3}{\vdSize}{Vicuna~1.3~13B Llama~160M} & Pretrained   &1.93\textsuperscript{$\pm$0.02}& $1.00 $&1.68\textsuperscript{$\pm$0.02}& $1.00 $&2.08\textsuperscript{$\pm$0.03}& $1.00 $&1.83\textsuperscript{$\pm$0.02}& $1.00 $&1.90\textsuperscript{$\pm$0.02}& $1.00 $&1.88\textsuperscript{$\pm$0.02}& $1.00 $ \\ 
& Distilled   &2.90\textsuperscript{$\pm$0.04}& $1.53 $&2.50\textsuperscript{$\pm$0.00}& $1.53 $&2.13\textsuperscript{$\pm$0.04}& $0.97 $&2.50\textsuperscript{$\pm$0.03}& $1.39 $&2.22\textsuperscript{$\pm$0.01}& $1.19 $&2.45\textsuperscript{$\pm$0.02}& $1.32 $ \\ 
& SD²   & \textbf{3.45\textsuperscript{$\pm$0.06}}& $\mathbf{1.83}$& \textbf{3.19\textsuperscript{$\pm$0.08}}& $\mathbf{1.96}$& \textbf{2.46\textsuperscript{$\pm$0.02}}& $\mathbf{1.14}$& \textbf{2.99\textsuperscript{$\pm$0.03}}& $\mathbf{1.67}$& \textbf{2.72\textsuperscript{$\pm$0.03}}& $\mathbf{1.46}$& \textbf{2.96\textsuperscript{$\pm$0.04}}& $\mathbf{1.61}$ \\ 
\midrule 
\multirow{3}{\vdSize}{Qwen3~14B Qwen3~0.6B} & Pretrained   &3.09\textsuperscript{$\pm$0.03}& $1.00 $&4.89\textsuperscript{$\pm$0.07}& $1.00 $&3.14\textsuperscript{$\pm$0.04}& $1.00 $&2.86\textsuperscript{$\pm$0.04}& $1.00 $&5.33\textsuperscript{$\pm$0.08}& $1.00 $&3.86\textsuperscript{$\pm$0.05}& $1.00 $ \\ 
& Distilled   &3.59\textsuperscript{$\pm$0.05}& $1.20 $&4.88\textsuperscript{$\pm$0.06}& $1.01 $&3.09\textsuperscript{$\pm$0.07}& $1.02 $&3.14\textsuperscript{$\pm$0.05}& $1.12 $&5.16\textsuperscript{$\pm$0.09}& $0.98 $&3.97\textsuperscript{$\pm$0.06}& $1.05 $ \\ 
& SD²   & \textbf{3.87\textsuperscript{$\pm$0.02}}& $\mathbf{1.28}$& \textbf{5.25\textsuperscript{$\pm$0.14}}& $\mathbf{1.08}$& \textbf{3.39\textsuperscript{$\pm$0.03}}& $\mathbf{1.10}$& \textbf{3.39\textsuperscript{$\pm$0.05}}& $\mathbf{1.19}$& \textbf{5.40\textsuperscript{$\pm$0.07}}& $\mathbf{1.01}$& \textbf{4.26\textsuperscript{$\pm$0.06}}& $\mathbf{1.11}$ \\ 
\midrule 
\multirow{3}{\vdSize}{Qwen3~8B Qwen3~0.6B} & Pretrained   &3.17\textsuperscript{$\pm$0.07}& $1.00 $& \textbf{5.18\textsuperscript{$\pm$0.09}}& $\mathbf{1.00}$&3.19\textsuperscript{$\pm$0.03}& $1.00 $&3.02\textsuperscript{$\pm$0.04}& $1.00 $&5.30\textsuperscript{$\pm$0.01}& $\mathbf{1.00}$&3.97\textsuperscript{$\pm$0.05}& $1.00 $ \\ 
& Distilled   &3.71\textsuperscript{$\pm$0.04}& $1.18 $&5.10\textsuperscript{$\pm$0.14}& $0.99 $&3.16\textsuperscript{$\pm$0.02}& $0.98 $&3.20\textsuperscript{$\pm$0.03}& $1.06 $&5.16\textsuperscript{$\pm$0.06}& $0.98 $&4.07\textsuperscript{$\pm$0.06}& $1.03 $ \\ 
& SD²   & \textbf{3.96\textsuperscript{$\pm$0.05}}& $\mathbf{1.24}$&5.18\textsuperscript{$\pm$0.07}& $0.99 $& \textbf{3.40\textsuperscript{$\pm$0.02}}& $\mathbf{1.05}$& \textbf{3.54\textsuperscript{$\pm$0.07}}& $\mathbf{1.16}$& \textbf{5.31\textsuperscript{$\pm$0.11}}& $0.99 $& \textbf{4.28\textsuperscript{$\pm$0.06}}& $\mathbf{1.06}$ \\ 
\midrule 
\multirow{3}{\vdSize}{Llama~3.1~8B Llama~3.2~1B} & Pretrained   &4.44\textsuperscript{$\pm$0.03}& $1.00 $&6.43\textsuperscript{$\pm$0.07}& $\mathbf{1.00}$&3.96\textsuperscript{$\pm$0.05}& $1.00 $&4.11\textsuperscript{$\pm$0.16}& $1.00 $& \textbf{5.62\textsuperscript{$\pm$0.08}}& $\mathbf{1.00}$&4.91\textsuperscript{$\pm$0.08}& $1.00 $ \\ 
& Distilled   &4.58\textsuperscript{$\pm$0.03}& $1.03 $&6.25\textsuperscript{$\pm$0.07}& $0.97 $&3.76\textsuperscript{$\pm$0.02}& $0.94 $&4.07\textsuperscript{$\pm$0.02}& $0.99 $&5.22\textsuperscript{$\pm$0.05}& $0.93 $&4.78\textsuperscript{$\pm$0.04}& $0.97 $ \\ 
& SD²   & \textbf{4.79\textsuperscript{$\pm$0.09}}& $\mathbf{1.07}$& \textbf{6.49\textsuperscript{$\pm$0.11}}& $0.99 $& \textbf{4.07\textsuperscript{$\pm$0.05}}& $\mathbf{1.02}$& \textbf{4.22\textsuperscript{$\pm$0.08}}& $\mathbf{1.02}$&5.44\textsuperscript{$\pm$0.06}& $0.95 $& \textbf{5.00\textsuperscript{$\pm$0.08}}& $\mathbf{1.00}$ \\ 
\midrule & \multicolumn{12}{c}{T=0 (Greedy)} \\
\midrule 
\multirow{3}{\vdSize}{Vicuna~1.3~13B Llama~160M} & Pretrained   &2.47& $1.00 $&2.08& $1.00 $&2.58& $\mathbf{1.00}$&2.26& $1.00 $&2.40& $1.00 $&2.36& $1.00 $ \\ 
& Distilled   &3.35& $1.39 $&3.01& $1.48 $&2.45& $0.92 $&2.86& $1.30 $&2.64& $1.12 $&2.86& $1.24 $ \\ 
& SD²   & \textbf{3.83}& $\mathbf{1.59}$& \textbf{3.63}& $\mathbf{1.80}$& \textbf{2.62}& $0.99 $& \textbf{3.26}& $\mathbf{1.48}$& \textbf{3.03}& $\mathbf{1.28}$& \textbf{3.27}& $\mathbf{1.43}$ \\ 
\midrule
\multirow{3}{\vdSize}{Qwen3~14B Qwen3~0.6B} & Pretrained   &3.13& $1.00 $&5.17& $1.00 $&3.30& $1.00 $&2.98& $1.00 $&5.57& $1.00 $&4.03& $1.00 $ \\ 
& Distilled   &3.82& $1.26 $&5.12& $1.00 $&3.30& $1.03 $&3.35& $1.14 $&5.45& $0.98 $&4.21& $1.06 $ \\ 
& SD²   & \textbf{4.05}& $\mathbf{1.33}$& \textbf{5.47}& $\mathbf{1.05}$& \textbf{3.61}& $\mathbf{1.11}$& \textbf{3.64}& $\mathbf{1.23}$& \textbf{5.66}& $\mathbf{1.01}$& \textbf{4.49}& $\mathbf{1.12}$ \\ 
\midrule 
\multirow{3}{\vdSize}{Qwen3~8B Qwen3~0.6B} & Pretrained   &3.24& $1.00 $&5.35& $1.00 $&3.38& $1.00 $&3.20& $1.00 $&5.41& $1.00 $&4.12& $1.00 $ \\ 
& Distilled   &3.93& $1.23 $&5.44& $\mathbf{1.02}$&3.46& $1.03 $&3.55& $1.12 $&5.47& $1.01 $&4.37& $1.07 $ \\ 
& SD²   & \textbf{4.15}& $\mathbf{1.28}$& \textbf{5.51}& $1.02 $& \textbf{3.71}& $\mathbf{1.09}$& \textbf{3.73}& $\mathbf{1.16}$& \textbf{5.58}& $\mathbf{1.02}$& \textbf{4.54}& $\mathbf{1.09}$ \\ 
\midrule 
\multirow{3}{\vdSize}{Llama~3.1~8B Llama~3.2~1B} & Pretrained   &4.51& $1.00 $&6.73& $\mathbf{1.00}$&4.11& $1.00 $&4.41& $1.00 $&5.86& $\mathbf{1.00}$&5.12& $1.00 $ \\ 
& Distilled   &4.89& $1.10 $&6.68& $1.00 $&4.06& $1.00 $&4.33& $0.98 $&5.82& $0.99 $&5.16& $1.01 $ \\ 
& SD²   & \textbf{5.04}& $\mathbf{1.11}$& \textbf{6.78}& $0.99 $& \textbf{4.24}& $\mathbf{1.03}$& \textbf{4.55}& $\mathbf{1.01}$& \textbf{5.93}& $0.99 $& \textbf{5.31}& $\mathbf{1.02}$ \\


\bottomrule
\end{tabularx}
\caption{Block efficiency and speedup across a variety of tasks for $k=8$, where UltraChat serves as the held-out validation set of training data for \SD and distilled. We report the block efficiency $\tau$ ($\pm$ denotes standard deviation over three independent evaluation runs; we further discuss statistical significance in \appref{app:stat_significance}) and speedup $\alpha$ over pretrained drafters for each verifier/drafter combination, ordered by decreasing drafter-to-verifier size ratio. Particularly for smaller pretrained drafters, as in the example of Vicuna~1.3, incorporating steering mechanisms significantly enhances throughput, achieving on average 61\% greater throughput and a 57\% increase in block efficiency compared to its pretrained counterpart under standard sampling. 
Llama~3.1's pretrained drafter already demonstrates higher block efficiency overall (0.94 higher block efficiency than Qwen3~8B \& Qwen3 0.6B on average), suggesting a naturally strong alignment between drafter and verifier. 
While distillation generally degrades performance across tasks not seen during training, \SD consistently preserves it.
For Qwen and Llama models, both distillation and \SD fail to improve over pretrained drafters already well-aligned on GSM8K and HumanEval datasets. Notably, \SD always achieves higher block efficiency than distillation and consistently achieves greater throughput.}
\label{table:results} 
\end{table*}

\subsection{Verification}
In \SD, the verification of candidate tokens remains unchanged to the regular speculative decoding framework, where we compute $\pi_V(\hat{x}_{t+i}\mid x_{1:t}, \hat{x}_{t+1:t+i-1})$ for all ${i \in [1, k]}$ in parallel, and then compare with the drafter's predicted outputs to accept or reject the token using rejection sampling, which has been proven to be optimal \cite{leviathan2023fast, yin2024theoretical}. We further enhance this step with the generation of a \emph{steering vector} $g_t$ to further condition $\pi_D$. This steering vector is generated based on the verifier's hidden states at the position of the first rejected token, i.e, the last token returned. Similar to \mbox{EAGLE-3} \cite{li2025eagle3} we use a linear layer, which is applied on the concatenation of $h_t, m_t, l_t$, which are the high, middle and low activations of the verifier from three different layers, to generate steering vector $g_t = W_{hml}[h_t, m_t, l_t]^\top$. 
\subsection{Drafting}
The drafting process of candidate tokens follows the standard auto-regressive decoding of regular LLMs, except that in \SD the probability distributions ${\pi_D(\hat{x}_{t+i} \mid x_{1:t}, g_t, \hat{x}_{t+1:t+i-1})}$ is further conditioned on the steering vector $g_t$. To steer the drafter, we incorporate a linear mapping of $g_t$ as a bias in all MLP layers $l=1,...,L_D$ of $\pi_D$ by changing the SwiGLU \cite{shazeer2020glu} computation from 
\begin{equation}
    a^{(l)}_{t+i} \mapsto W_d(W_ua^{(l)}_{t+i}  \odot \sigma(W_g a^{(l)}_{t+i} )),
    \label{eq:before_steering}
\end{equation}
to 
\begin{equation}
    a^{(l)}_{t+i}, \mathbf{g_t} \mapsto W_d((W_ua^{(l)}_{t+i} + \mathbf{W_s g_t})  \odot \sigma(W_g a^{(l)}_{t+i} )). 
    \label{eq:after_steering}
\end{equation}
This ensures that the added overhead is negligible compared to the latency of the transformer, while allowing for a large amount of control in all layers of the drafter, as now the MLP is not solely conditioned on the hidden state, but also the steering vector. As $W_s g_t$ is invariant to drafting position $i$, we can compute it once at the beginning of each drafting stage. Note that the steering vector also influences the keys/ values in the attention mechanism, meaning the computation of $\pi_D(\hat{x}_{t+i} \mid x_{1:t}, g_t, \hat{x}_{t+1:t+i-1})$ is not only conditioned on $g_t$, but also all prior steering vectors $g_{t^\prime}$ used in prior tokens.
\begin{figure*}[t]
    \centering
    \includegraphics[width=0.9\textwidth]{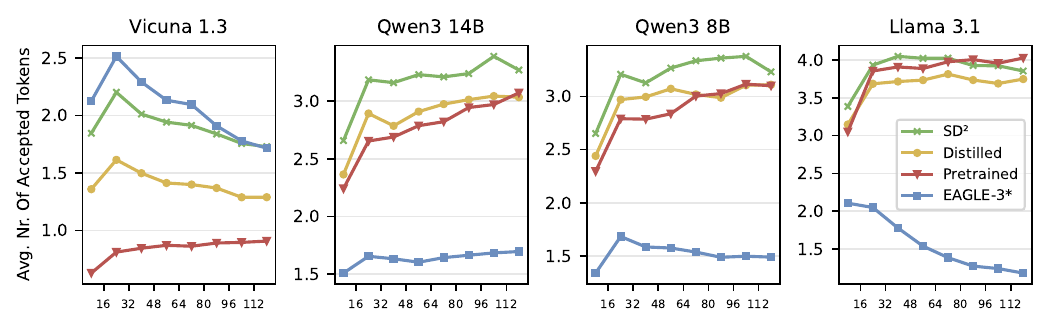}
    \caption{Number of tokens accepted per block at different positions. We compare how different drafter/verifier pairs fare at different positions throughout the generation process: A point at position $x$ means the average number of accepted tokens per block for blocks with the last generated token having position $x\pm8$. As can be seen, large pretrained drafters can leverage their vast training data to maintain strong drafting performance with increased sequence length. \SD minimally interferes with this behavior. }
    \label{fig:loss_over_time}
\end{figure*}

\subsection{Training}
To train \SD we utilize synthetic data generated by $\pi_\text{V}$ and use the probability distribution $\pi_V(x_t|x_{1:t-1})$ as targets. 
Similar to \citet{zhou2024distillspec}, we use a synthetic dataset, as they have shown better alignment improvements compared to ground truth data, as it better reflects the verifier's behavior at inference.
To train the steering mechanism, we utilize a uniformly random offset $\delta \in [1,k]$ and compute $\pi_D(x_t|x_{1:t-1},g_{t-\delta})$. This ensures that the steering mechanism uniformly receives gradients for all drafting positions and hence must learn to encode information about the upcoming $k$ tokens. In addition to the steering mechanism, we also fully fine-tune the drafter, while the verifier remains frozen throughout training to ensure lossless acceleration. This step is critical to the performance improvement of \SD, as observed in our ablation presented in \cref{fig:ablation_all}.
\citet{leviathan2023fast} showed that total variational distance ($D_\text{TVD}$) is equivalent to the rejection rate, making it the natural choice as a criterion. However, \citet{zhou2024distillspec} showed that the choice of loss is more nuanced and showed that Kullback–Leibler divergence ($D_\text{KL}$),  which we adopt, often outperforms $D_\text{TVD}$ as a criterion. 
The initialization of the steering mechanism is crucial, as too much interference by the untrained mechanism can lead to the model diverging. We initialize $W_s =0$ and $W_{hml}$ such that $W_{hml}\space [h_t, m_t , l_t]^\top = h_t + m_t + l_t$.

\section{Experiments}

\paragraph{Baselines.} We evaluate the efficacy of our method against several drafting strategies: \emph{Pretrained}, which employs speculative decoding with the unchanged drafter and \emph{Distilled} \cite{zhou2024distillspec}, which first aligns the drafter to the verifier at training time. Additionally, we evaluate EAGLE-3\textsuperscript{*} \cite{li2025eagle3}, a state-of-the-art dependent drafter used as a baseline for block efficiency. The asterisk indicates that we restrict EAGLE-3 to chain drafting mode to ensure a fair comparison and consistency with the other models.

\paragraph{Model Configurations.}
To assess the performance of \SD, we use 4 different open source verifier-drafter pairs: 
\emph{Vicuna~1.3} 13B with Llama 160M, \emph{Qwen3~14B} and Qwen3 0.6B, \emph{Qwen3~8B} and Qwen3 0.6B, and \emph{Llama~3.1} 8B‐Instruct and Llama~3.2 1B‐Instruct. \cite{zheng2023judging, miao2024specinfer, yang2025qwen3, meta2024llama3.1, meta2024llama3.2}
These configurations were selected to represent a range of verifier–drafter capacity gaps and model families. 
For EAGLE-3\textsuperscript{*}, we use the publicly released weights trained on UltraChat and ShareGPT datasets \cite{li2025eagle3}.

\paragraph{Tasks.}
We run experiments on 96 samples from 5 different datasets: The held-out validation split of \emph{UltraChat\_200k} \cite{ding2023enhancing} for dialogue, \emph{HumanEval} \cite{chen2021evaluating} for code generation, \emph{XSum} \cite{narayan2018dont} for summarization, \emph{Alpaca} \cite{dubois2024lengthcontrolled} for instruction-following, and \emph{GSM8K}  \cite{cobbe2021training} for reasoning. These common datasets provide coverage across core capabilities such as reasoning, summarization, and interaction. From this list, UltraChat\_200k is the only dataset that the distilled and \SD drafters have seen during training. 

\paragraph{Decoding Parameters.}
We fix the drafter’s draft length to $k=8$ tokens, yielding speculation blocks of size $k+1=9$. All decoding is performed using the chain drafting strategy. We consider two sampling regimes: full sampling with temperature $T=1$, and greedy decoding with $T=0$. We use batched speculative decoding with a batch size of 12 and generate up to 128 output tokens per example. 
To ensure statistical reliability under stochastic sampling, we generate outputs at $T=1$ across three distinct random seeds and report their mean and standard deviation. Due to memory limitations, we limit the total number of tokens computed (including rejected ones) to 512 tokens. For Vicuna~1.3, we relax this constraint by reducing the batch size and disabling the maximum token count, due to the low acceptance rate of the pretrained model. 

\paragraph{Metrics.}
Since speculative decoding preserves the base model’s probability distribution, this study focuses solely on efficiency metrics: \emph{Block efficiency ($\tau$)} and \emph{speedup compared to the pretrained independent drafter ($\alpha$)}. Block efficiency refers to the tokens generated per block, and is a driving factor in the efficacy of speculative decoding. This metric can be derived from the number of accepted tokens per block, and adding 1 for the token generated from the joint drafter-verifier distribution after rejection. We measure speedup as the increase in tokens generated per second compared to the independent pretrained drafter. Note that speedup is hardware-dependent, unlike hardware-agnostic metrics such as $\tau$. 

\paragraph{Training Details.}
The distilled and \SD drafters are initialized from the pretrained drafter and finetuned for 6 epochs on synthetic data generated by the verifier with temperature $T=1$, using prompts sourced from UltraChat\_200k \cite{ding2023enhancing}, limited to a total sequence length of 256. Training is conducted with an effective batch size of 24 using the AdamW \cite{loshchilov2019decoupled} optimizer. Refer to \appref{app:training_details} for more info.
After training on UltraChat, we fine-tune each drafter for one more epoch on synthetic samples derived from the ShareGPT dataset \cite{sharegpt2023}. All experiments are performed on one NVIDIA A100 GPU with 80GB of memory.
\begin{figure}[t]
    \centering
    \includegraphics[width=0.9\linewidth]{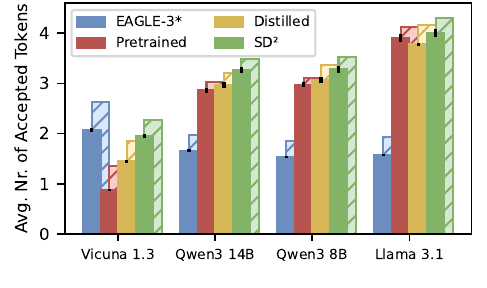}
    \caption{The average number of accepted tokens per Block for the different speculative decoding setups  (Left to right: EAGLE-3\textsuperscript{*}, Pretrained, Distilled, \SD) averaged across all tasks. Solid bars correspond to $T=1$ (sampling), and the hashed bars to $T=0$ (greedy). One can see that \SD consistently achieves higher acceptance rates compared to both the Distilled and Pretrained drafter. In Vicuna~1.3, the number of active parameters for the drafter (Llama 160M) is less than half as many as the respective EAGLE-3\textsuperscript{*} model. At such small sizes, pretrained drafters lose their competitiveness to dependent heads; however, \SD can bridge this gap.}
    \label{fig:n_accepted}
\end{figure}

\subsection{Ablations}
\begin{figure}[t]
    \centering
    \includegraphics[width=0.89\linewidth]{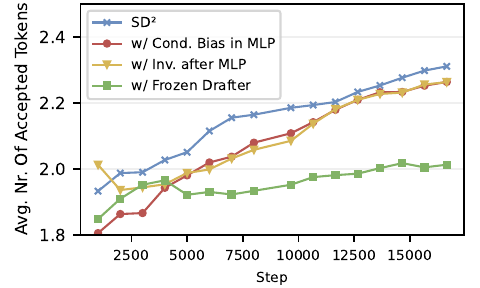}
    \caption{Number of tokens accepted for ablation experiments throughout training for 2 epochs on Vicuna~1.3 7B ($\pi_V$) and Llama 160M ($\pi_D$).
    We omit step 0, where every experiment has the value of the pretrained drafter, of 1.0.
    \SD utilizes a bias in the MLP, right after the up-projection, and unfreezes the drafter. \emph{Inv. Bias after MLP} simplifies the steering mechanism of \SD by adding the bias right after the MLP, while \emph{Cond. Bias in MLP} increases the modeling capability of the steering mechanism by instead calculating the bias based on not only $g_{t}$, but also $\smash{f^{(l-1)}_{t+i}}$. We also test keeping the drafter frozen during training. 
    \SD consistently outperforms the other variants, justifying our design choices.
    }
    \label{fig:ablation_all}
\end{figure}
We justify the design of the steering mechanism and training in \SD with ablation studies on \emph{Vicuna-1.3 7B} and \emph{Llama 160M}. We investigate two key \SD design choices: The steering mechanism and unfreezing the drafter. Further ablation studies can be found in \appref{app:ablation}.

\paragraph{What steering mechanism is most effective?}
The choice of a steering mechanism, i.e, how we modify the behavior of the drafter conditioned on the steering vector $g_t$, is critical for the effectiveness of \SD. We aim to design an optimal steering mechanism that (i) is latency-lightweight, (ii) remains compatible with other acceleration techniques like KV-Caching, and (iii) provides precise control over the drafter’s behavior. We evaluated three options that differ in the number of parameters and expressiveness. The simplest method adds a bias $W_s g_t$ right after the MLP for all hidden layers, so $\smash{\tilde{f}_{t+i}^{(l)}:=f_{t+i}^{(l)} + W_s g_t}$ for all $i \in [1,k]$. Note that $W_s g_t$ only has to be computed once, as it is invariant of draft position $i$. The second approach, which we ultimately adopt in \SD (see~\cref{eq:after_steering}), modifies this by instead conditioning the existing MLP on $g_t$ for all layers by adding $W_s g_t$ to the up-projection, right before gating. The last approach modifies all layers by adding a 2-layer MLP with input $\smash{f_{t}^{(l-1)}}$ and $g_t$, which computes the bias, which in turn is then used inside the MLP as in the second approach. As seen in~\cref{fig:ablation_all}, \SD consistently scores the highest block efficiency. 

\paragraph{Should one fine-tune the drafter?}
In \SD, we fine-tune both the steering mechanism and the drafter parameters. To isolate the effect of steering, we also evaluate a frozen-drafter variant where only the steering is trained. 
As inferrable from~\cref{fig:ablation_all}, steering alone can increase the number of tokens accepted by +100\% over the unaligned pretrained's baseline of 1.0, indicating that the verifier’s hidden states convey valuable guidance and can meaningfully influence the drafter’s output. 
However, as shown in \cref{fig:ablation_all}, unfreezing the drafter consistently yields better results.

\subsection{Results}
\definecolor{rejRed}{cmyk}{0, 0.67, 0.74, 0.26}
\definecolor{accGreen}{cmyk}{0.76, 0, 0.76, 0.53} 
\definecolor{finBlue}{cmyk}{0.63, 0.47, 0, 0.39}
\newcommand{\acc}[1]{\textcolor{accGreen}{#1}}
\newcommand{\rej}[1]{\textcolor{rejRed}{\st{#1}}}
\newcommand{\fin}[1]{\textcolor{finBlue}{\underline{#1}}}
\begin{table*}[t]
\centering
{\small
\begin{tabularx}{\linewidth}{XXp{5.7cm}}
\toprule
\textbf{Pretrained} & \textbf{Distilled} & \textbf{\SD (ours)}  
\\ \midrule
\rej{Here is the implemented Python function `has} \fin{To}  \acc{solve this problem, we need to} \rej{ **} \fin{ determine}  \acc{ if} \rej{ there exists at least two elements in} \fin{ **} \acc{any} \rej{ pair of numbers in a list is} \fin{ two} \acc{numbers} \rej{ in a list** are **clo} \fin{**} ...
& 
\rej{Here's the implemented Python function `has} \fin{To}  \acc{ solve this problem, you need to} \rej{:\textbackslash n\textbackslash n} \fin{ determine}  \acc{ whether} \rej{ there exists any two numbers in the} \fin{ **} \acc{ any} \rej{ pair of numbers in a list is} \fin{ two}  \acc{ numbers} \rej{ in the list** differ}... 
&
\acc{To solve the problem}\rej{, we need to} \fin{ of}  \acc{ determining whether any two numbers in a list} \fin{ are}  \acc{ closer to each other than a given `}\fin{threshold} \acc{`, we can} \rej{ use a **hash map} \fin{ approach}  \acc{ it} \rej{ as follows:\textbackslash n\textbackslash n\#\#\# [?][?] Approach}...
    \end{tabularx}}
    \caption{Qualitative example of speculative decoding with different drafting methods. \emph{\acc{Green tokens}} are accepted, \emph{\rej{red tokens}} are rejected, and the \emph{\fin{blue token}} is the final token per block sampled from the joint drafter-verifier distribution. Note that the symbol [?] refers to tokens outside of the English alphabet, highlighting the inherent risk of hidden state intervention. Continuation and more examples can be found in \appref{app:sd_example}.}
    \label{table:sd_example}
\end{table*}

\paragraph{Block Efficiency} 
\cref{table:results} and \cref{fig:n_accepted} summarize the results across all configurations. \SD consistently yields a higher block efficiency compared to both distilled and pretrained approaches. This improvement is particularly pronounced on UltraChat, the evaluation set from the training distribution, where \SD shows clear advantages. Across all datasets, \SD either matches or surpasses the performance of the pretrained model. 
As can be seen in \cref{fig:n_accepted}, the Qwen and especially the Llama pretrained drafters already achieve high acceptance rates. This is particularly true in GSM8K and HumanEval, as can be seen in \cref{table:results}, where the distilled drafter is consistently outmatched by the pretrained drafter. This suggests that the distilled version has overfit to tasks in the style of UltraChat dialogue. While \SD also degrades in performance on these tasks compared to UltraChat or similar tasks, it can consistently match or beat both pretrained and distilled drafters.
In the case of Vicuna~1.3, the pretrained drafter is not closely aligned to the model. This is expected, as unlike other drafter-verifier pairs, these models were trained on different data distributions and have a larger capacity gap. In that setting, as seen in \cref{table:results}, both distillation and \SD significantly increase the block efficiency. For instance, with $T{=}1$ sampling, the distilled drafter achieves an average block efficiency improvement of 0.57 over the pretrained baseline across all tasks, while \SD further adds 0.51 accepted tokens per block.
As seen in \cref{fig:n_accepted}, both distilled and \SD perform reliably under both $T=0$ (greedy decoding) and $T=1$ (sampling), demonstrating robustness to different decoding regimes. On average, the integration of steering leads to an increase of 21\% on the number of tokens accepted per block ($\tau -1$) compared to distilled drafters and 31\% compared to pretrained ones. 

\paragraph{Performance in Long Sequence Drafting.}
A key advantage of using pretrained drafters is their exposure to large-scale datasets and long-context training, which equips them with strong generation capabilities over extended sequences. As demonstrated in \cref{fig:loss_over_time}, both distilled and \SD maintain this capability. \SD consistently has more accepted tokens, and therefore also higher block efficiency, compared to both the pretrained and the distilled drafter across a range of token positions. Moreover, as evident by \cref{fig:loss_over_time}, \SD maintains a relatively constant advantage over distillation across all token positions, showing that steering works well with increasing sequence length. A continuation of the example in \cref{table:sd_example} is available in \appref{app:sd_example}.

\paragraph{Speedup over Pretrained.}
\cref{table:results} shows that, despite adding a small amount of computational latency to the drafting operation, \SD can speed up pretrained models by up to +83\% on training data, while distilled models achieve an improvement of up to +53\% over baseline. Across all tasks in Vicuna~1.3, \SD achieves a speedup of +61\% under regular sampling and +43\% under greedy sampling. On average, \SD provides a speedup of +19.5\% for $T=0$ and +16.3\% for $T=1$ compared to its pretrained counterpart. Crucially, \SD achieves roughly twice the additive speedup compared to distilled drafting under standard sampling and roughly 75\% more additive speedup with greedy sampling. Furthermore, for Qwen3~8B on HumanEval with $T{=}1$, we observe that the steered method incurs only a 1\% slowdown while matching the block efficiency of independent drafting, confirming the mechanism’s minimal overhead.

\section{Limitations and Future Work}
The performance of \SD, much like distillation-based approaches, is highly dependent on the composition and quality of the training data. Although \SD often matches the pretrained drafter on out-of-domain tasks, its effectiveness remains strongest on data similar to its training distribution, as shown in \cref{table:results}. This highlights the importance of either training on a comprehensive and diverse dataset or limiting the drafter to a singular domain. 
Furthermore, while achieving higher block efficiency, it provides little to no speedup over an already well-aligned drafter, such as Llama~3.1. 
Moreover, changing hidden representations in transformer networks is a delicate matter, as small changes in the wrong direction, as evidenced in \cref{table:sd_example}, can lead the model to produce nonsensical output.
Additionally, while speculative decoding with pretrained drafters can isolate the verifier in a black box, \SD requires access to the verifier's hidden states. This can be challenging in applications involving external remote verifiers.
While we demonstrate that steering can be retrofitted onto existing drafters, we do not explore the training of new drafters explicitly designed for dynamic steering, which we leave as an open direction for future work. 
Furthermore, we do not compare \SD’s steering to more invasive methods like EAGLE’s concatenation of verifier states. However, \SD’s key advantage is its modularity, as steering can be added post hoc without requiring verifier signals during pretraining.
All models in this study use SwiGLU \cite{shazeer2020glu} in their feedforward layers. While our steering mechanism should generalize to other gated activations, this remains to be validated in future work.
Finally, extending \SD to more complex speculative decoding paradigms, such as dynamic tree verification, remains an open problem, with application-specific studies needed to assess its practical viability and competitiveness against other speculative decoding paradigms.
\section{Conclusion}
This study presents a method to dynamically steer pretrained drafters during speculative decoding, achieving substantial performance improvements compared to baselines and across a wide range of drafter and verifier configurations. 
In addition to improving acceptance rates, our system exhibits greater robustness on out-of-distribution tasks, suggesting that steering mechanisms are less susceptible to over-fitting on the training task. 

\section*{Acknowledgments}
We thank Benjamin Estermann for his valuable input and discussions during the early stages of this project.
    
{\small
\bibliography{references}
}

\ifisextended
    \appendix
    \include{appendix}
\else
\fi

\end{document}

%% file: appendix.tex
\onecolumn
\section{Technical Appendix}

\subsection{Notation}
We use the following notation throughout the paper:
\begin{itemize}
    \item $\pi_V$: Verifier model (also referred to as the base model).
    \item $\pi_D$: Drafter model.
    \item $k$: Number of tokens generated (drafted) per step.
    \item $\tau$: Block efficiency metric.
    \item $\alpha$: Speed up in throughput compared to using the pretrained drafter.
    \item $x_{i:j} := \{x_k \mid i \leq k \leq j\}$: Token subsequence from position $i$ to $j$, inclusive.
    \item $g_t$: Steering vector generated by $\pi_V$ at timestep $t$.
    \item $h_t, m_t, l_t$: High-, middle-, and low-level hidden states from $\pi_V$ at timestep $t$.
    \item $f_t^{(l)}$: Hidden state at timestep $t$ after the $l$-th layer.
    \item $T = 0, T = 1$: Sampling temperatures; $T = 1$ corresponds to standard sampling, and $T = 0$ to greedy sampling.
    \item $[a, b] := \{i \in \mathbb{N} \mid a \leq i \leq b\}$: Closed interval over natural numbers from $a$ to $b$.
\end{itemize}

\subsection{Training \& Initialization Details}\label{app:training_details}
Training was done on 1xA100-80GB with BF-16 precision. The drafter was stored in full precision throughout training. We use a cosine learning rate scheduler with 1000 warm-up steps starting and ending at 1/10th the learning rate. We employ gradient-norm clipping with a value of 0.5 to ensure stability. For AdamW we set $\beta_1 = 0.9$, $\beta_2 = 0.999$, and model-specific learning rates (\mbox{5e-5} for Vicuna~1.3, \mbox{1e-5} for Llama~3.1 and \mbox{2e-5} for Qwen3~8B~\&~14B). Regarding initialization, we set $W_s$ as the $0$ matrix, which means that the drafter's behavior is initially unchanged by the noisy steering vector. We add a layer norm right after the computation of the steering vector, to stabilize the steering mechanism. We select the $h,m,l$ layers inspired by EAGLE-2 choosing layers $3, L/2, L-2$ 

\subsection{Ablation Studies}
\label{app:ablation}
\begin{figure}[ht]
     \centering
     \includegraphics[width=0.5\linewidth]{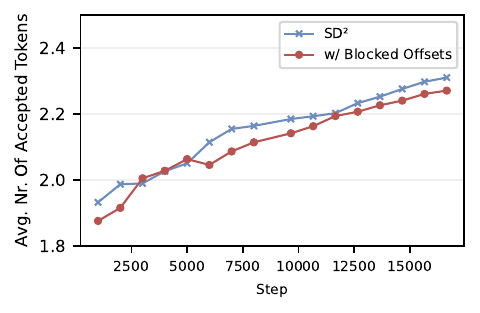}
     \vspace{-1em}
     \caption{\textbf{Block efficiency throughout training with different training offset methods}: The training process of \SD uses a constant offset, i.e $g_{t}$ is used for the prediction of $\hat{y}_{t+\delta}$, for training, where $\delta \in [1,k]$ random. We compare with \emph{blocked offsets}, as they mimic the behavior observed during inference, by using $g_t$ for the prediction of $\hat{y}_{t+1:t+k}$ if $t \operatorname{mod} k =\delta-1$. \SD slightly beats the blocked offset method. Here, the verifier used is Vicuna 1.3 7B, and Llama 160M is the drafter. }
     \label{fig:ablation_offset}
 \end{figure}

 \paragraph{Should one model the offsets used in training after inference?}\label{app:ablation_offset}
 During training, our method samples a random offset $\delta \in [1, k]$ and uses the steering vector $g_t$ to predict the distribution of $\hat{x}_{t+\delta}$. This differs from inference, where $g_t$ conditions the prediction of all tokens $\hat{x}_{t+1:t+k}$. To better simulate inference behavior during training, we instead apply a \textit{blocked offset} strategy: $g_t$ is used as the steering vector for the entire prediction block $\hat{y}_{t+1:t+k}$, where $t \operatorname{mod} k=\delta-1$. As can be seen in \cref{fig:ablation_offset}, these two methods only differ slightly, with \SD having the slight edge. 

\subsection{Statistical Significance}
We evaluate the statistical significance of our results using a pair-wise Welch's t-test across datasets and drafter-verifier pairs in \cref{table:statistics}.
\label{app:stat_significance}

\begin{table*}[ht]  
\centering  
\small
\renewcommand{\arraystretch}{0.8}
\begin{tabularx}{\linewidth}{lll XXXXXXXXXX} 
\toprule
\textbf{Verifier} & \textbf{Drafter} & \textbf{Method} & \multicolumn{2}{c}{\textbf{UltraChat}} & \multicolumn{2}{c}{\textbf{HumanEval}} & \multicolumn{2}{c}{\textbf{XSum}} & \multicolumn{2}{c}{\textbf{Alpaca}} & \multicolumn{2}{c}{\textbf{GSM8K}} \\ 
 & & & $t$ & $p$  & $t$ & $p$ & $t$ & $p$ & $t$ & $p$ & $t$ & $p$ \\

\midrule 
\multirow{2}{*}{Vicuna 1.3 13B} & \multirow{2}{*}{Llama 160M} & Distilled   &40.15&\textbf{0.00}&72.90&\textbf{0.00}&1.82&0.16&35.19&\textbf{0.00}&32.51&\textbf{0.00} \\ 
& & SD²   &43.70&\textbf{0.00}&32.96&\textbf{0.00}&18.65&\textbf{0.00}&54.96&\textbf{0.00}&47.49&\textbf{0.00} \\ 
\midrule 
\multirow{2}{*}{Qwen3 14B} & \multirow{2}{*}{Qwen3 0.6B} & Distilled   &13.91&\textbf{0.00}&-0.26&0.81&-1.04&0.37&8.19&\textbf{0.00}&-2.41&0.07 \\ 
& & SD²   &35.40&\textbf{0.00}&3.88&\textbf{0.03}&9.20&\textbf{0.00}&14.56&\textbf{0.00}&1.19&0.30 \\ 
\midrule 
\multirow{2}{*}{Qwen3 8B} & \multirow{2}{*}{Qwen3 0.6B} & Distilled   &11.43&\textbf{0.00}&-0.83&0.46&-1.54&0.20&6.12&\textbf{0.00}&-4.03&0.05 \\ 
& & SD²   &15.99&\textbf{0.00}&-0.05&0.96&9.80&\textbf{0.00}&11.94&\textbf{0.00}&0.24&0.83 \\ 
\midrule 
\multirow{2}{*}{Llama 3.1 8B} & \multirow{2}{*}{Llama 3.2 1B} & Distilled   &6.35&\textbf{0.00}&-3.14&\textbf{0.03}&-6.28&\textbf{0.02}&-0.48&0.68&-7.60&\textbf{0.00} \\ 
& & SD²   &6.55&\textbf{0.01}&0.85&0.45&2.56&0.06&1.06&0.37&-3.25&\textbf{0.04} \\ 

\bottomrule
\end{tabularx}
\caption{Pairwise t-tests on block efficiency for $T=1$ across 3 distinct seeds. We evaluate the statistical significance of each of the results by considering a pair-wise t-test between the pretrained baseline and the distilled version or \SD. $t$ corresponds to Welch's t-statistic, which measures the standardized difference. A positive t-statistic indicates an improvement over the pretrained drafter. $p$ corresponds to the probability of the null hypothesis, i.e., that there is no difference between the two distributions. We marked all $p\leq0.05$ in bold. 
}
\label{table:statistics} 
\end{table*}

\subsection{Example of speculative decoding}
\label{app:sd_example}
See \cref{table:sd_example_cont} for a continuation of the example presented in \cref{table:sd_example}.
\begin{table}[ht]
\centering
\renewcommand{\arraystretch}{0.5}
\small
\begin{tabularx}{\linewidth}{p{5.4cm}Xp{5.9cm}}
\toprule
\textbf{Pretrained} & \textbf{Distilled} & \textbf{\SD (ours)}  
\\ \midrule
  \acc{} \rej{Here is the implemented Python function `has} \fin{To} \acc{solve this problem, we need to} \rej{**} \fin{determine} \acc{if} \rej{there exists at least two elements in} \fin{**} \acc{any} \rej{pair of numbers in a list is} \fin{two} \acc{numbers} \rej{in a list** are **clo} \fin{**} \acc{in the} \rej{list `numbers` are **} \fin{given} \acc{list} \rej{of floating-point numbers are **clo} \fin{are} \acc{} \rej{closer to each other than a specified **} \fin{**} \acc{closer to each other than the given} \rej{} \fin{threshold} \acc{**.\textbackslash{}n\textbackslash{}n\#\#\#} \rej{[?][?] Type hints (} \fin{Strategy} \acc{:\textbackslash{}n\textbackslash{}n- **} \rej{Check all pairs** of} \fin{Sort} \acc{the list** of numbers} \rej{.\textbackslash{}n- **} \fin{first} \acc{} \rej{.\textbackslash{}n- **Check adjacent pairs** in} \fin{.} \acc{} \rej{\textbackslash{}n- **Iterate through the sorted} \fin{When} \acc{} \rej{sorted, two numbers that are close to} \fin{the} \acc{list is sorted, the closest} \rej{elements to} \fin{numbers} \acc{} \rej{between two elements are at the **ends} \fin{will} \acc{be} \rej{near the ends of the list.\textbackslash{}n} \fin{**} \acc{adjacent**} \rej{in the list, and} \fin{.\textbackslash{}n} \acc{-} \rej{Use a **linear scan** to} \fin{Then} \acc{,} \rej{check if any pair of **adj} \fin{iterate} \acc{through the sorted list and compare} \rej{each pair} \fin{**} \acc{each} \rej{number with the previous one** to} \fin{pair} \acc{of adjacent numbers**} \rej{with the threshold.\textbackslash{}n} \fin{.\textbackslash{}n} \acc{- If} \rej{any such pair has elements that} \fin{the} \acc{} \rej{difference between them is **less than the} \fin{**} \acc{} \rej{difference between two consecutive elements is less than} \fin{absolute} \acc{difference} \rej{between any two adjacent numbers** is} \fin{**} \acc{between any two adjacent numbers} \rej{**less than} \fin{is} \acc{**less than the threshold**, return `} \rej{} \fin{True} \acc{`.\textbackslash{}n- If no such pair is found} \rej{} \fin{after}& \acc{} \rej{Here's the implemented Python function `has} \fin{To} \acc{solve this problem, you need to} \rej{:\textbackslash{}n\textbackslash{}n} \fin{determine} \acc{whether} \rej{there exists any two numbers in the} \fin{**} \acc{any} \rej{pair of numbers in a list is} \fin{two} \acc{numbers} \rej{in the list** differ by **} \fin{**} \acc{in the list are **closer} \rej{than} \fin{to} \acc{each other than the given threshold**.} \rej{} \fin{This} \acc{} \rej{is a **common problem** found in} \fin{can} \acc{be done} \rej{in a few key steps:\textbackslash{}n\textbackslash{}n} \fin{efficiently} \acc{} \rej{using a **hash map (dictionary)**} \fin{by} \acc{} \rej{checking all possible pairs of elements (i} \fin{**} \acc{} \rej{finding the minimum distance** between any two} \fin{sorting} \acc{the list**, and then checking} \rej{for elements} \fin{the} \acc{} \rej{pairs later based on their element and its} \fin{**} \acc{} \rej{distance between neighboring elements**, since saying that} \fin{difference} \acc{between each pair of consecutive elements**.\textbackslash{}n\textbackslash{}n} \rej{} \fin{\#\#\#} \acc{Key Steps:\textbackslash{}n\textbackslash{}n1. **Sort the} \rej{} \fin{list} \acc{**} \rej{so that the smallest number comes first} \fin{–} \acc{} \rej{This helps in easily finding neighboring elements.\textbackslash{}n} \fin{Sorting} \acc{} \rej{the array allows us to easily iterate through} \fin{makes} \acc{it easier to} \rej{analyze consecutive elements.\textbackslash{}n2} \fin{check} \acc{for adjacent elements, because the} \rej{list becomes} \fin{closest} \acc{elements will be next to each other} \rej{.\textbackslash{}n} \fin{in} \acc{} \rej{the sorted list.\textbackslash{}n\textbackslash{}n2. **Iter} \fin{a} \acc{sorted list.\textbackslash{}n2. **Check adjacent} \rej{} \fin{elements} \acc{** – For each adjacent pair in the} \rej{} \fin{sorted} \acc{list} \rej{(i.e., `sorted\_numbers} \fin{,} \acc{compute the absolute difference} \rej{between the two.\textbackslash{}n} \fin{.} \acc{If any} \rej{of these differences **less than} \fin{difference} \acc{is less than the} \rej{given threshold, return} \fin{threshold} \acc{, return `True`.\textbackslash{}n3. **} \rej{} \fin{Return}& \acc{To solve the problem} \rej{, we need to} \fin{of} \acc{determining whether any two numbers in a list} \rej{} \fin{are} \acc{closer to each other than a given `} \rej{} \fin{threshold} \acc{`, we can} \rej{use a **hash map} \fin{approach} \acc{it} \rej{as follows:\textbackslash{}n\textbackslash{}n\#\#\# [?][?] Approach} \fin{systematically} \acc{.\textbackslash{}n\textbackslash{}nThe} \rej{idea is to:\textbackslash{}n\textbackslash{}n1.} \fin{key} \acc{idea is} \rej{that we can:\textbackslash{}n\textbackslash{}n1.} \fin{:\textbackslash{}n\textbackslash{}n} \acc{-} \rej{Compare **all** pairs of distinct} \fin{Sort} \acc{the list of numbers} \rej{in ascending order.\textbackslash{}n} \fin{,} \acc{} \rej{which helps in efficiently comparing all pairs (} \fin{because} \acc{after sorting, the} \rej{indices of elements that} \fin{closest} \acc{} \rej{values to each other (i.e.,} \fin{pair} \acc{} \rej{will always be the second and second-to} \fin{of} \acc{numbers will be} \rej{consecutive elements in the sorted} \fin{adjacent} \acc{} \rej{in the original list.\textbackslash{}n- Traverse the} \fin{.\textbackslash{}n} \acc{-} \rej{For each element in the sorted list} \fin{Then} \acc{, iterate through the sorted list and check} \rej{} \fin{if} \acc{the difference between} \rej{the current number and the} \fin{any} \acc{two adjacent numbers is less than the} \rej{given} \fin{`} \acc{threshold`.\textbackslash{}n\textbackslash{}nHere's the implementation} \rej{in Python} \fin{of} \acc{the function} \rej{you provided, with a bit} \fin{with} \acc{the correct} \rej{`threshold`, using `typing} \fin{logic} \acc{:\textbackslash{}n\textbackslash{}n```python\textbackslash{}nfrom typing import List} \rej{} \fin{\textbackslash{}n\textbackslash{}n} \acc{def has\_close\_elements(numbers: List[float} \rej{} \fin{],} \acc{threshold: float) -\textgreater bool:\textbackslash{}n} \rej{} \fin{"""} \acc{Check if in given list of numbers,} \rej{} \fin{any} \acc{two numbers are closer to each other\textbackslash{}n} \rej{} \fin{} \acc{than} \rej{given threshold.\textbackslash{}n    \textbackslash{}n    \textgreater\textgreater\textgreater has} \fin{the} \acc{given threshold.\textbackslash{}n\textbackslash{}n    \textgreater\textgreater\textgreater has\_close\_elements} \rej{} \fin{([} \acc{} \rej{than given threshold.\textbackslash{}n    \textbackslash{}n    \textgreater\textgreater\textgreater has} \fin{float} \acc{} \rej{'given threshold.\textbackslash{}n    \textgreater\textgreater\textgreater has\_close} \fin{\textbackslash{}n} 
 \end{tabularx}
\caption{Continuation of qualitative example shown in \cref{table:sd_example}. The text is generated using different drafting methods with $T=1$ sampling and the Qwen3 14B verifier/drafter setup. \emph{\acc{Green tokens}} are accepted, \emph{\rej{red tokens}} are rejected, and the \emph{\fin{blue token}} is the final token per block sampled from the joint drafter-verifier distribution.}
\vspace{-9em}
\label{table:sd_example_cont}
\end{table}